\documentclass[10pt,article]{IEEEtran}
\usepackage[pdftex]{graphicx}
\usepackage{dblfloatfix}
\usepackage{multirow}
\usepackage{changes}
\usepackage{longtable}
\usepackage{graphicx} 
\usepackage{subcaption} 
\usepackage{tabularx}
\pdfpageattr{/Group << /S /Transparency /I true /CS /DeviceRGB >>}

\usepackage[utf8]{inputenc}
\usepackage[T1]{fontenc}
\usepackage{textcomp}
\usepackage{microtype}
\usepackage{calc}
\usepackage[normalem]{ulem}

\usepackage{lipsum}

\usepackage[range-phrase=--,per-mode=symbol-or-fraction,binary-units=true,range-units=single,list-units=single,detect-all,forbid-literal-units=true]{siunitx}
\AtBeginDocument{
	\DeclareSIUnit\bit{bit}
	\DeclareSIUnit\byte{Byte}
	\DeclareSIUnit\decibelm{dBm}
	\DeclareSIUnit\vehicle{veh}
}

\usepackage{silence}
\usepackage{csquotes}
\usepackage[backend=biber,style=ieee,doi=false,isbn=false,mincitenames=1,maxcitenames=2]{biblatex}
\DeclareFieldFormat{sentencecase}{#1} % never apply sentence casing, even if bibtex field is unprotected
\DeclareFieldFormat{titlecase}{#1} % never apply title casing, even if bibtex field is unprotected
\usepackage{xpatch}
\xpatchbibmacro{textcite}{\addspace}{\addnbspace}{}{}
\xpatchbibmacro{Textcite}{\addspace}{\addnbspace}{}{}
\DefineBibliographyStrings{english}{
	andothers = et~al\adddot\addspace
}

\DeclareSourcemap{
	\maps[datatype=bibtex]{
		\map[overwrite=true]{
			\step[fieldsource=school,fieldtarget=institution]
		}
	}
}

\usepackage{amsmath}
\usepackage{amssymb}
\usepackage{amsfonts}
\usepackage{amsthm}

\usepackage{booktabs}
\usepackage{url}

\usepackage[inline]{enumitem}

\usepackage[caption=false,font=footnotesize]{subfig}
\usepackage[pdftex]{graphicx}
\usepackage{dblfloatfix}
\DeclareGraphicsExtensions{.pdf,.png,.jpg}
\usepackage{tikz}
\usetikzlibrary{arrows}
\usetikzlibrary{calc}
\usetikzlibrary{chains}
\usetikzlibrary{scopes}

\usepackage[american]{babel}
\usepackage{hyphenat}

\usepackage{algorithmic}
\usepackage{algorithm}

\usepackage[capitalize,noabbrev,nameinlink]{cleveref}

\RequirePackage{xstring}
\RequirePackage{xparse}
\RequirePackage[]{acro}
\makeatletter
\@ifpackagelater{acro}{2019/02/17}{
	\NewDocumentCommand\acrodef{mO{#1}mG{}}{\DeclareAcronym{#1}{short={#2}, long={#3}, foreign-plural={}, #4}}
}{
	\NewDocumentCommand\acrodef{mO{#1}mG{}}{\DeclareAcronym{#1}{short={#2}, long={#3}, #4}}
}
\makeatother
\usepackage[color]{changebar}
\def\todoCtd#1{%
	TODO: #1%
	\ifx&#1&...\fi%
	\endgroup
	\cbend
	\relax
}

\NewDocumentCommand\IEEE{ s m >{\SplitArgument{4}{/}}d[] }{%
	\IfBooleanTF{#1}{}{IEEE\,}% suppress IEEE when using starred form
	\nolinebreak[2]% this is a somewhat bad place for a line break
	#2%
	\IfNoValueTF{#3}{%
	}{%
		\sommerIEEELettersSlashed#3%
	}%
}
\newcommand{\sommerIEEELettersSlashed}[5]{%
	\IfNoValueTF{#2}{%
	}{%
		\nolinebreak[3]% multiple letters, this is just a very bad place for a line break
	}%
	#1%
	\IfNoValueTF{#2}{}{/#2}%
	\IfNoValueTF{#3}{}{/#3}%
	\IfNoValueTF{#4}{}{/#4}%
	\IfNoValueTF{#5}{}{/#5}%
}

\bibliography{references.bib}

\acrodef{ML}{Machine Learning}
\acrodef{LLMs}{Large Language Model}
\acrodef{FNN}{Feedforward Neural Network}
\acrodef{LLaMA}{Large Language Model Meta AI}
\acrodef{RMSQLN}{Root Mean Square Layer Normalization}
\acrodef{GQA}{Grouped-Query Attention}
\acrodef{RoPE}{Rotary Position Embedding}
\acrodef{SwiGLU}{Swish-Gated Linear Unit}
\acrodef{NLP}{Natural Language Processing}

\begin{document}

%\title{An Evaluation of Large Language Models for the Detection of Political Orientation in Newspapers}
\title{Large Language Models' Detection of Political Orientation in Newspapers}

% SUGGESTED JOURNALS
%- PNAS (https://www.pnas.org/) TOO EXPENSIVE (maybe ask for waiver?)
%- PNAS Nexus (https://academic.oup.com/pnasnexus) [not the same thing] TOO EXPENSIVE
%- Journal of the Association for Information Science and Technology *
%- Nature Machine Intelligence *
%- International Journal of Data Science and Analytics **
%- IEEE Transactions on Computational Social Systems ***
%- Transactions of the Association for Computational Linguistics (https://transacl.org/index.php/tacl/about) ****

\author{%
\IEEEauthorblockN{%
	Alessio Buscemi\IEEEauthorrefmark{1}, 
        Daniele Proverbio\IEEEauthorrefmark{2}
}

\IEEEauthorblockA{
	\IEEEauthorrefmark{2}Department of Industrial Engineering, University of Trento, Via Sommarive, 9, 38123, Povo (TN), Italy
}%

\texttt{%
    \IEEEauthorrefmark{1}alessio.buscemi0208@gmail.com
    \IEEEauthorrefmark{2}daniele.proverbio@unitn.it
}%
}

%%% DP affiliation: Department of Industrial Engineering, University of Trento, 9 via Sommarive, 38123 Trento

% \IEEEpubid{\begin{minipage}{\textwidth}\ \\[12pt] \centering
%   1551-3203 \copyright 2015 IEEE. Personal use is permitted, but republication/redistribution requires IEEE permission.\\
%   See http://www.ieee.org/publications standards/publications/rights/index.html for more information.
% \end{minipage}} 

% -------------- Section end marker --------------
%                _       _
%               ( )_    ( )
%    ___  _   _ | ,_)   | |__     __   _ __   __
%  /'___)( ) ( )| |     |  _ `\ /'__`\( '__)/'__`\
% ( (___ | (_) || |_    | | | |(  ___/| |  (  ___/
% `\____)`\___/'`\__)   (_) (_)`\____)(_)  `\____)
%
% -------------- Section end marker --------------

\maketitle

\begin{abstract}%
  Democratic opinion-forming may be manipulated if newspapers' alignment to political or economical orientation is ambiguous. Various methods have been developed to better understand newspapers' positioning. Recently, the advent of Large Language Models (LLM), and particularly the pre-trained LLM chatbots like ChatGPT or Gemini, hold disruptive potential to assist researchers and citizens alike. However, little is know on whether LLM assessment is trustworthy: do single LLM agrees with experts' assessment, and do different LLMs answer consistently with one another? In this paper, we address specifically the second challenge. We compare how four widely employed LLMs rate the positioning of newspapers, and compare if their answers align with one another. We observe that this is not the case. Over a woldwide dataset, articles in newspapers are positioned strikingly differently by single LLMs, hinting to inconsistent training or excessive randomness in the algorithms. We thus raise a warning when deciding which tools to use, and we call for better training and algorithm development, to cover such significant gap in a highly sensitive matter for democracy and societies worldwide. We also call for community engagement in benchmark evaluation, through our open initiative \texttt{navai.pro}.
\end{abstract}
\begin{IEEEkeywords}
Journalism, ChatGPT, Gemini, Large Language Models, Artificial Intelligence
\end{IEEEkeywords}

\acresetall
\IEEEpeerreviewmaketitle

% -------------- Section end marker --------------
%                _       _
%               ( )_    ( )
%    ___  _   _ | ,_)   | |__     __   _ __   __
%  /'___)( ) ( )| |     |  _ `\ /'__`\( '__)/'__`\
% ( (___ | (_) || |_    | | | |(  ___/| |  (  ___/
% `\____)`\___/'`\__)   (_) (_)`\____)(_)  `\____)
%
% -------------- Section end marker --------------

%Use the acronym package to insert acronyms, such as \ac{ACC}.
%This is the first time the \ac{ECU} acronym is used and this is the second time \acp{ECU}.
%The bandwidth is \SI{100}{\mega\bit\per\second} and %\SI{50}{\kilo\meter\per\hour}, \SI{2}{\square\kilo\meter}.
%\numlist{2;3;4;5}.
%\numrange{4}{8}.

\section{Introduction}
\label{sec:intro}

Newspapers -- printed or online -- are paramount elements of the information media ecosystem, and the orientation of people's information is often shaped by journalists' writing -- articles, reports, comments or editorials \cite{adelman2000death, beck2012grundlagen}. Traditionally, newspapers lean towards specific positioning in the economical and political spectrum, following agendas and promoting specific points of views, for selected target audiences \cite{gentzkow2014competition, calvano2021market, spinde2020enabling}. We thus witness an array of newspapers styles and positioning, from "generalist" newspapers, roughly positioned around centrist coordinates, to left- or right-winged prints which tend to align to their respective political agendas, often promoting similarly oriented economical standpoints. Different positioning is often accompanied by use of certain wording, and keywords, tones and other textual features may contribute to the identification of editorial leaning \cite{levendusky2013partisan, reah2002language}. 

Correctly identifying newspapers' political and economical alignment is crucial to better interpret their messages, uncover potentially hidden agendas and biases, and correlate with trustworthiness \cite{falck2018sentiment, lin2023high, cage2020media}. As the contemporary news reporting ecosystem is obscured by infodemics, misinformation and other subtle nuances of non-objective information reporting \cite{wardle2017information,lewandowsky2020technology}, improving the detection of newspaper positioning would thus provide valuable information to researchers, practitioners and citizens. Fact-checking and bias identification, which are paramount to promote democratic debates, rely on correct and timely detection of news media alignment \cite{graves2013deciding}. 

Databases listing the average positioning of the major newspapers are often updated \cite{ansolabehere2006orientation, barclay2014indian}, but they lack at identifying the orientation of single articles, or at tracking the orientation dynamics over time. In addition, they may not be fully accessible to citizens. Alteratively, scraping through articles to extract meaningful and dynamical information is time-consuming and often requires expert supervision. Hence, automated pipelines for sentiment or positioning analysis have been developed over the years, from the Political Compass initiative \cite{politicalcompass} to sentiment analysis-based studies \cite{enevoldsen2017analysing}, to data-driven initiatives \cite{falck2018sentiment, congleton2022tracing} (see also \cite{spinde2023media} for a recent review). Political orientation and alignment in textual content refers to the ideological positioning conveyed by the language, which can range from conservative to liberal or from authoritarian to libertarian, among others. Interpreting these orientations involves detecting biases, sentiments, and underlying values expressed in the text.

The recent skyrocketing of Transformers-based deep learning models may provide a breakthrough in digital humanities in general \cite{suissa2022text}, and in positioning classification in particular \cite{spinde2023media,spinde2022exploiting, raza2024dbias}. Among the most recent tools, Large Language Models (LLMs) are particularly well-suited for this task due to their deep understanding of language semantics and syntax, learned from training on diverse and expansive text corpora. Boasting state-of-the-art capabilities in text mining \cite{beltagy2019scibert}, natural language processing \cite{min2023recent} and sentiment analysis \cite{buscemi2024chatgpt}, LLMs are powerful instruments that researchers and journalists may successfully use to unlock new avenues for mapping and understanding the world news media landscape. Among the numerous LLMs currently available, pre-trained LLMs, which are often embedded in popular chatbots like Chat-GPT by OpenAI \cite{chatgpt}, LlaMa by Meta \cite{touvron2023llama}, Gemini by Google \cite{team2023gemini} or Claude by Anthropic \cite{claude}, hold impressive disruption potential due to their computation power and pervasiveness into society, as well as the investments that support their development and adoption by academicians and the broader public. \\

A key question to address before adopting such models for studies and industrial applications is thus: do they work properly? Here, we intend the specification "properly" in two meanings. First, whether different LLMs work consistently with each other, that is, if the same article from the same journal is classified similarly by distinct LLMs and if, by analyzing multiple articles, there are conserved patterns among LLMs. Second, whether the positioning that is automatically generated conforms with some "ground truth", assessed by experts in the field. While the second point obviously refers to the ability of LLMs to provide valid answers, the first point relates to the inherent opaqueness of such models, which encompass extremely complex architectures that are trained on different datasets to fit billions of parameters. This results in the impossibility to assess \textit{a priori} whether the answers should be expected to hold consistency. Evaluating the responses \textit{post-hoc} and cross-checking across LLMs allows to address this point, probing whether they found consistent and repeated patterns in the data or if they return randomic or unbearably volatile answers.

Here, we mainly address the first point, subsequently hinting at the second one. After making a collection of newspapers worldwide, we select articles over time using an automated scraper, to reproduce research methodologies and make an agnostic selection of articles to test. Then, we process them using state-of-the-art pre-trained LLMs, employing default settings that professionals and citizens may likely employ. This way, we verify whether LLMs are consistent in their mapping of media positioning and assess the implications of their usage in global societies.

\section{Background}
\label{sec:background}

This section provides background knowledge regarding polarization and journalism and LLMs.

\subsection{Polarization of journalism}
\label{sub:polarization}

The digital age comes with intensified ideological polarization, posing several risks to journalism and democratic societies, like:

\begin{itemize}
    \item \textbf{Echo Chambers and Filter Bubbles}: Algorithms on social networks and search engines promote filter bubbles, limiting exposure to diverse viewpoints and fostering environments where only like-minded perspectives are shared. This isolation reinforces existing beliefs and deepens ideological divides \cite{sunstein2017republic, cinelli2021echo}.
    \item \textbf{Decline of Local News}: The financial struggles of traditional media have led to the decline of local news outlets. This reduction in local journalism diminishes community engagement and oversight, allowing national and international issues to overshadow local concerns \cite{abernathy2018expanding, heese2022local}.
    \item \textbf{Partisan Media}: The fragmentation of the media landscape has given rise to explicitly partisan news outlets. These outlets often prioritize sensationalism and ideological alignment over balanced reporting, contributing to a polarized public discourse \cite{levendusky2013partisan}.
    \item \textbf{Erosion of Trust}: The blurring of lines between news, opinions, and entertainment, coupled with the spread of fake news, has eroded public trust in media. This skepticism can lead to disengagement from mainstream news sources and increased reliance on alternative, often less reliable, information channels \cite{guess2021consequences}. %\cite{newman2020reuters, }
\end{itemize}

The 2016 and 2020 presidential elections highlighted the extent of media polarization in the U.S. Social media platforms were used to disseminate targeted misinformation campaigns, which contributed to deepening political divides. The role of partisan news outlets, such as Fox News and MSNBC, further exacerbated this polarization \cite{benkler2018network}.

In regions with emerging digital infrastructures, such as parts of Africa and Asia, the rapid adoption of social media has outpaced the development of robust journalistic standards and fact-checking mechanisms. This situation has made these regions particularly vulnerable to misinformation and polarizing content \cite{wasserman2018global}.

\subsection{AI for the evaluation of political orientations}
\label{sub:ai_for_evaluation}

The rapid development of Large Language Models (LLMs) has opened new avenues for analyzing and interpreting vast amounts of textual data. One significant application of LLMs is in identifying the political alignment of news articles \cite{rony2024exploring, leite2023detecting}. This capability is crucial in an era where media bias and misinformation are pervasive concerns. By leveraging the nuanced understanding of language that LLMs possess, researchers aim to classify news content as left-leaning, right-leaning, or centrist, providing valuable insights into media landscapes.

LLMs exhibit remarkable proficiency in natural language understanding, enabling them to detect subtle cues and patterns indicative of political bias. These models can analyze the language, tone, and framing of news articles to infer political orientation. Key capabilities include \cite{zhang2023revisiting,wei2022emergent}:

\begin{itemize}
    \item \textit{Contextual Analysis}: LLMs can understand the context in which specific terms and phrases are used, allowing them to differentiate between neutral and politically charged language.
    \item \textit{Sentiment Detection}: By assessing sentiment, LLMs can infer the underlying bias, such as positive or negative sentiments towards political figures or policies.
    \item \textit{Topic Modeling}: LLMs can identify and categorize topics within an article, helping to correlate certain subjects with known political biases.
    \item \textit{Pattern Recognition}: The models can recognize patterns in writing styles and rhetoric that are characteristic of particular political orientations.
\end{itemize}

These capabilities make LLMs powerful tools for identifying political orientation, as they can process and analyze data at a scale and speed beyond human capacity. 

Despite their strengths, LLMs face several challenges in accurately identifying the political orientation of news articles, which can be summarised as:

\begin{itemize}
    \item \textit{Training Data Bias}: LLMs are trained on large datasets that may contain inherent biases. If the training data is skewed towards certain political perspectives, the model's predictions may reflect those biases.
    \item \textit{Ambiguity and Nuance}: Political orientation is not always clear-cut. Articles may contain mixed signals or subtle biases that are challenging to detect. LLMs may struggle with ambiguity and the nuanced nature of political discourse.
    \item \textit{Evolving Language}: Political language and terminology evolve over time. LLMs may require continuous updating to stay current with new expressions and shifts in political rhetoric.
    \item \textit{Cultural and Regional Differences}: Political orientation can vary significantly across different cultures and regions. An LLM trained primarily on Western media, for instance, may not accurately assess articles from other parts of the world.
    \item \textit{Overfitting to Patterns}: LLMs might overfit to specific patterns seen in the training data, leading to incorrect generalizations when encountering new or less common expressions of political orientation.
\end{itemize}

\section{Methods}
\label{sub:methodology}

This section outlines the methodology used for collecting and processing the articles for the models, as well as the choice of LLM models and their querying. The entire project was implemented using Python 3.10 and LLM-specific APIs.

\subsection{Investigation pipeline}
The dataset to test the responses of LLMs was designed to include multiple articles from newspapers around the globe, featuring different languages. The pipeline to collect and process articles that defined the dataset is illustrated in \cref{alg1}.

First, we initialize an empty list $E$ (step 1), subsequently populated with the evaluations performed by the LLMs. Then, we run through a list $U$ collecting homepages of target websites. $U$ was constructed to include a relevant number of mostly-read newspapers from around the world, in various languages. Each URL $u$ in the list $U$ is processed independently (steps 2-9). \\

For this study, we used the widely-known \texttt{Beautiful Soup4} library as the scraper. The scraper identifies up to $N$ unique hyperlinks from $u$ (step 3). They serve as candidate URLs, potentially containing the articles to be evaluated by the LLMs.

A subset of size $S < N$ from the scraped URLs is then selected (step 4). This subset is determined from the longest URLs, based on the observation that short links typically lead to category pages (e.g., 'Sport', 'Economy', etc.) rather than proper articles.

Articles are then processed from the selected URLs using a second scraper, namely \texttt{newspaper}, which specializes in newspaper articles (step 5). A subset of these scraped articles is selected based on their length, between user-defined \textit{MIN} and \textit{MAX} limits (step 6). The reasons for this selection are:

\begin{itemize}
\item If an article is too short, it might not be a proper article (flash-news, sponsored content etc.) or may contain insufficient information for the model to assess the political inclination.
\item If an article is too long, it can significantly impact the model's evaluation capability (as discussed in \cref{sub:models}).
\item The length of articles can greatly affect the costs incurred in querying proprietary models (see section \cref{sub:models}).
\end{itemize}

The selected articles are then individually passed to the LLMs (listed in \cref{tab:parameters} and discussed in \cref{sub:models}) for evaluation (step 7). 

When $A$ articles per day have been correctly processed (i.e., no errors occurred and the output is in the correct format), the evaluations are appended to $E$ (step 8). \\

\begin{algorithm}[htb]
	\begin{algorithmic}[1]
	\REQUIRE List of homepage URLs $U$, Maximum number of URLs to scrape from homepage $N$, Number of selected URLs $S$, Minimum length of articles \textit{MIN}, Maximum length of articles \textit{MAX}, Number of articles to evaluate per newspaper $A$, Model $M$
	\ENSURE Evaluations $E$
    \STATE $E$ $\gets$ ()
    \FOR {$u$ \textbf{in} $U$}
        \STATE \textit{scraped\_urls} $\gets$ scrape\_urls($u$, $N$)
        \STATE \textit{selected\_urls} $\gets$ select\_urls(\textit{scraped\_urls}, $S$) 
        \STATE \textit{articles} $\gets$ extract\_articles(\textit{selected\_urls})
        \STATE \textit{selected\_articles}$\gets$ select\_articles(\textit{articles}, \textit{MIN}, \textit{MAX})
        \STATE \textit{evals} $\gets$ query($M$, \textit{selected\_articles},  $A$)
        \STATE $E$.append(\textit{evals})
    \ENDFOR
	\end{algorithmic}
 \caption{Article selection algorithm.}
    \label{alg1}
\end{algorithm}

% \section{Performance evaluation}
% \label{sub:performance}

% In this section, we present the dataset employed for the performance evaluation of our study, and we discuss the results obtained from our analysis.

% \subsection{Set up}
% \label{sub:parameters}

\subsection{Parameter setting}

The newspapers list was initially constructed by pooling 152 newspapers (see Appendix \ref{app:newspapers}), sourced from various media outlets to ensure a diverse and representative sample. Out of those newspapers, several had to be discarded, and 40 were eventually selected and analysed. The primary criteria for their selection were the success of the automated scraping process and the availability of a sufficient number of articles meeting the \textit{MIN} and \textit{MAX} requirements. These factors were pivotal in determining the feasibility of consistent data extraction and subsequent analysis, thereby shaping the composition of our final dataset. As a result, $U=40$ homepage URLs, each corresponding to a newspaper, have been scraped. 

The complete list is reported in \cref{tab:newspapers}. The selected newspapers span 27 countries across the globe and encompass a wide range of political alignment, positioning themselves over the full spectrum of possibilities without predominant clusters. Based on literature and web sources, they are categorised as follows: 5 right-winged, 10 centre-right, 5 centre, 6 centre-left, 4 left, and 10 that are either independent or unknown. The political alignment often correlates with economical orientation. This diverse geographical and positioning representation ensures a wide array of perspectives and contexts. 

\begin{table*}[htbp]
\centering
\caption{Selected newspapers, homepage URL and positioning, according to web and literature sources. For newspapers marked with $^*$, it was not possible to reliably verify the positioning or political alignment through sources.}
\begin{tabular}{lllll}
\hline
\textbf{Country} & \textbf{Newspaper} & \textbf{Homepage} & Positioning & Source \\
\hline
ARG & Clarín & \url{https://www.clarin.com} & Centre-right & https://www.lifeder.com/periodicos-izquierda-derecha/ \\
AUT & Kurier & \url{https://kurier.at} & Centre-right & \cite{sandford2013encyclopedia} \\
BEL & De Morgen & \url{https://www.demorgen.be} & Left & https://www.demorgen.be/\\
BEL & De Standaard & \url{https://www.standaard.be} & Centre-right & \cite{standaard} \\
BEL & Le Soir & \url{https://www.lesoir.be} & Independent & \cite{standaard} \\
BRA & Estadão & \url{https://www.estadao.com.br} & Right & \cite{carvalho2013imprensa} \\
BRA & O Globo & \url{https://oglobo.globo.com} & Right & \cite{shahin2016protesting} \\
BRA & The Rio Times  & \url{https://riotimesonline.com} & Independent & https://riotimesonline.com \\
CAN & La Presse & \url{https://www.lapresse.ca} & Centre & https://www.lapresse.ca \\
CZE & Lidové Noviny & \url{https://www.lidovky.cz} & Centre-right & \cite{gawrecka2013watches} \\
DEU & Die Zeit & \url{https://www.zeit.de} & Centre-left & \cite{hess2009german} \\
FRA & Le Figaro & \url{https://www.lefigaro.fr} & Centre-right & Enc. Britannica: "Le Figaro"\\
FRA & Le Monde & \url{https://www.lemonde.fr} & Centre-left & lemonde.fr \\
FRA & Libération & \url{https://www.liberation.fr} & Left & liberation.fr \\
GRC & Kathimerini & \url{https://www.kathimerini.gr} & Centre & Bloomberg: "Kath. Pub. SA - Company Profile and News" \\
HUN & 24.hu & \url{https://24.hu} & Centre & 24.hu \\
IND & LiveMint & \url{https://www.livemint.com} & -$^*$ & - \\
IND & The Hindu & \url{https://www.thehindu.com} & Centre-left & thehindu.com \\
IND & The Times of India & \url{https://timesofindia.indiatimes.com} & -$^{*}$ & - \\
ISR & The Jerusalem Post & \url{https://www.jpost.com} & Right & \cite{dridi2020reporting} \\
ITA & Libero Quotidiano & \url{https://www.liberoquotidiano.it} & Centre-right & liberoquotidiano.it \\
LBN & The Daily Star & \url{http://www.dailystar.com.lb} & -$^*$ & - \\
LUX & Le Quotidien & \url{https://lequotidien.lu} & Left & Euro Topics: "The media landscape in Luxembourg" \\
LUX & Tageblatt & \url{https://www.tageblatt.lu} & Centre-left & Euro Topics: "The media landscape in Luxembourg" \\
MYS & The Star & \url{https://www.thestar.com.my} & Centre-right & \cite{hilley2001malaysia} \\
MEX & El Universal & \url{https://www.eluniversal.com.mx} & Right & Media Ownership Monitor Mexico 2018 \\
MEX & La Jornada & \url{https://www.jornada.com.mx} & Left &  Media Ownership Monitor Mexico 2018  \\
NLD & De Volkskrant & \url{https://www.volkskrant.nl} & Centre & volkskrant.nl \\
NGA & Punch & \url{https://punchng.com} & -$^*$ & - \\
NOR & Aftenposten & \url{https://www.aftenposten.no} & Independent & \cite{cook2014europe}\\
PAK & Dawn & \url{https://www.dawn.com} & -$^*$ & - \\
PER & El Comercio & \url{https://elcomercio.pe} & Centre-right & \cite{barnhurst1993layout} \\
QAT & Al Sharq & \url{https://www.al-sharq.com} & -$^*$ & - \\
ESP & La Razón & \url{https://www.larazon.es} & Centre-right & \cite{numbers2014creationism} \\
TUR & Hürriyet & \url{https://www.hurriyet.com.tr} & Centre & \cite{ozyurek2006nostalgia} \\
USA & Chicago Tribune & \url{https://www.chicagotribune.com} & Centre-right & Boston University Library WR150 \\
USA & Fox News & \url{https://www.foxnews.com} & Right & \cite{dellavigna2007fox} \\
USA & The New York Times & \url{https://www.nytimes.com} & Centre-left & Boston University Library WR150 \\
USA & Star Tribune & \url{https://www.startribune.com} & Centre-left & Media Bias Fact Check \\
URY & El País & \url{https://www.elpais.com.uy} & -$^*$ & - \\
\hline
\end{tabular}
\label{tab:newspapers}
\end{table*}

Regarding $N$, it was set to 200 after observing that a newspaper homepage typically contains between 50 and 300 hyperlinks, many of which lead to categories, videos, etc. Preliminary tests indicated that 200 was a sufficiently large number to identify a sufficient number of articles that meet the eligibility criteria for evaluation later in the pipeline. $S$ was set at 20 to select at least the top 10\% of the longest URLs.

The article length was then set between 1000 and 5000 characters using \textit{MIN} and \textit{MAX}, corresponding to short-to-medium-length articles, which, based on our observations, constitute the majority of articles found in the selected newspapers. As explained above, articles of these lengths are ideal for evaluation through LLMs.

By selecting $A$ = 5 and processing a batch of articles per day for each newspaper over 5 consecutive days, from May 9th to May 13th, 2024, we obtained a test set of 1000 articles (40 newspapers $ \cdot$ 5 articles per day $ \cdot$ 5 days).

\cref{tab:parameters} summarises the values of the parameters chosen for the study.

\begin{table}[htbp]
\centering
\caption{Parameters chosen or the evaluation}
\begin{tabular}{lll}
\hline
\textbf{Parameter} & \textbf{Value} \\
\hline
List of homepage URLs $U$ & 40 \\
Max number of URLs to scrape from homepage $N$ &  200\\
Number of selected URLs $S$ & 20 \\
Minimum length of articles \textit{MIN} & 1000 chars \\
Maximum length of articles \textit{MAX} & 5000 chars\\
Number of articles to evaluate per newspaper $A$ & 5\\
Model $M$ & ChatGPT-4, \\
& Gemini Pro 1.5, \\
& ChatGPT-3.5,\\
& Gemini Pro\\
\hline
\end{tabular}
\label{tab:parameters}
\end{table}

\subsection{Models}
\label{sub:models}

The articles are evaluated using the most recent LLM-based chatbots ChatGPT-4, Gemini Pro 1.5, ChatGPT-3.5 and Gemini Pro.
According to the latest benchmark at the time of writing \cite{openai2024evaluation}, ChatGPT-4 and Gemini Pro 1.5 are the models with the highest capabilities. We also included their previous versions, ChatGPT-3.5 and Gemini Pro, to gain a comprehensive understanding of the differences and similarities arising from the evolution of these models. They are also still widely used and are easily accessible. This approach allows us to investigate how advancements in these models impact their performance and outputs.

ChatGPT-3.5 and ChatGPT-4 are LLMs from the series of the Generative Pre-trained Transformer (GPT) models developed by OpenAI \cite{chatgpt}. 
ChatGPT-3.5 features enhancements in language comprehension and response accuracy compared to its predecessor, GPT-3, due to refinements in training techniques and data handling. 
ChatGPT-4, the more sophisticated successor, further expands the model's capabilities with a significantly larger parameter count and improvements in training algorithms. 
This version demonstrates superior performance in a broader range of natural language understanding and generation tasks, showcasing enhanced contextual understanding, reduced biases, and a greater ability to handle nuanced and complex queries \cite{buscemi2024chatgpt}. 
Both models operate on the transformer architecture.

Gemini Pro 1.0 and 1.5 from Google AI \cite{team2023gemini} have been trained on a massive dataset of text and code. 
However, Pro 1.5 offers several advancements over Pro 1.0. First, while Pro 1.0 details are not publicly available, Pro 1.5 utilizes a multimodal mixture-of-experts architecture \cite{reid2024gemini}. With this, it efficiently handles different data types (text, code, audio, video) within a single framework. Moreover, Pro 1.5 is considered a mid-sized model. While smaller than some prior models, it is claimed to achieve similar performance to the larger Gemini 1.0 Ultra \cite{hassabis2024}. Moreover, a key strength of Pro 1.5 is its extended context window: while Pro 1.0 likely has a smaller window, limiting its ability to reason over long sequences, Pro 1.5 boasts a default window of 1 million tokens, allowing it to process information equivalent to hours of audio/video, entire codebases, or lengthy documents. 
Finally, Pro 1.5 demonstrates improved performance on various benchmarks compared to Pro 1.0, particularly in long-document tasks \cite{reid2024gemini}.

These models are based on a token-based cost structure. Every token is an individual linguistic unit that the model processes, both as input and as output (usually input tokens are less expensive than output tokens). Between ChatGPT-4 and ChatGPT-3.5, we observed a significant cost difference; for our experiment, ChatGPT-4 was almost 20 times more expensive than ChatGPT-3.5. 
We also encountered certain restrictions when using Gemini Pro 1.5 throughout our testing. We employed its free version as the premium version was not yet accessible. However, because of the daily quota constraints, the testing requires increasing time delays with the length and number of articles.

\subsection{Query}

The goal of the present study is to check whether LLM rate news media positioning in a consistent manner with one another, and if they succeed in replicating the distribution of political alignment that emerges from our dataset. To this end, we employ a commonly used compass framework to represent the political and economical spectrum \cite{heywood2021political}. This maps political and economical alignment onto a horizontal axis for socio-economic factors and a vertical axis for socio-cultural considerations. The model thus grades news articles on numerical scales, according to their alignment towards economic left/right and libertarian/authoritarian positions. \\

To obtain the desired mapping, all LLM models have been prompted with the same query, to minimise bias associated with the use different prompts. The query is:

\textit{Instructions: Economic Scale from -10 to 10, where -10 is Economic Left and 10 is Economic Right. Scale Democracy from -10 to 10, where -10 is Libertarian and 10 is Authoritarian. I provide a newspaper article. Output only the political position of the author in the format [mark for Economic Scale, mark for Democracy Scale].
NEVER write any text before or after the result.
ALWAYS provide the result, even if you are not fully sure.
Article: \{Article\}} 

The query formulation was chosen after a meticulous iterative process of prompt engineering, including trial and error to maximize the likelihood of obtaining the desired output. Notably, assertive language with short instructions was used. The words 'never' and 'always' were capitalized because, based on information from the models themselves, they interpret capitalized text as more imperative.

We intentionally did not specify whether marks should be integers or decimals.  Interestingly, based on our observations, all LLM interpreted the required evaluation as integers.

Finally, note that we refer to the author's opinion as a concise way to say 'the opinion expressed by the author in this article'. In any case, the author's name is hidden from the model, which only processes the body of the article, ensuring that LLMs have no prior biases regarding the author's identity.

\section{Results}
\label{sub:results}

Multiple articles, over 40 global newspapers, have been positioned on a political-economical compass, ranging from economic left/right and political libertarian/authoritarian positions (including the in-between spectrum). Do different LLMs return similar mapping? \\

\begin{figure*}[ht]
    \centering
    \begin{subfigure}[b]{0.42\linewidth}
        \includegraphics[width=\linewidth]{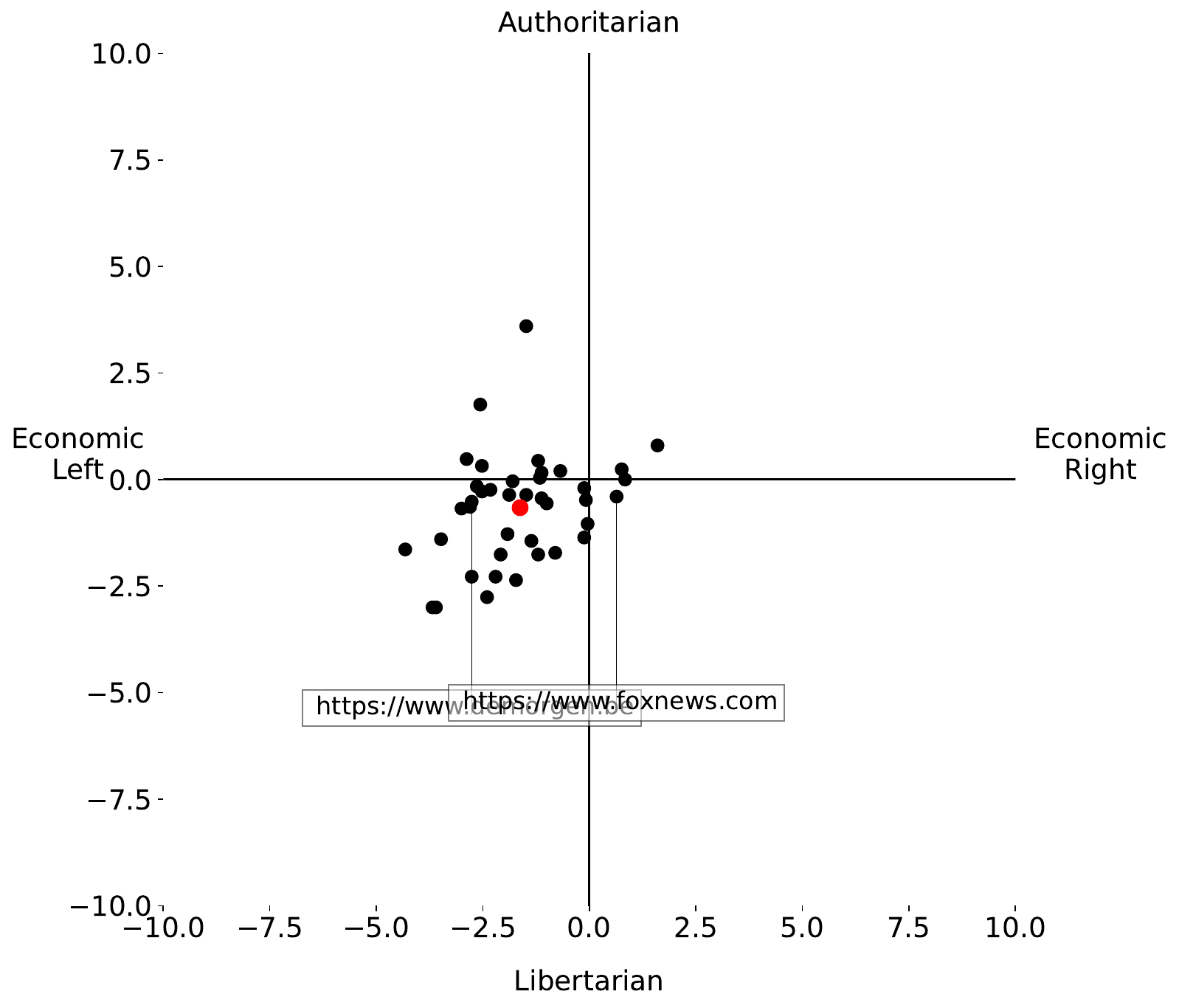}
        \caption{ChatGPT-3.5}
        \label{fig:sub1_1}
    \end{subfigure}
    %\hfill % this will add a little horizontal space between the subfigures
    \begin{subfigure}[b]{0.42\linewidth}
        \includegraphics[width=\linewidth]{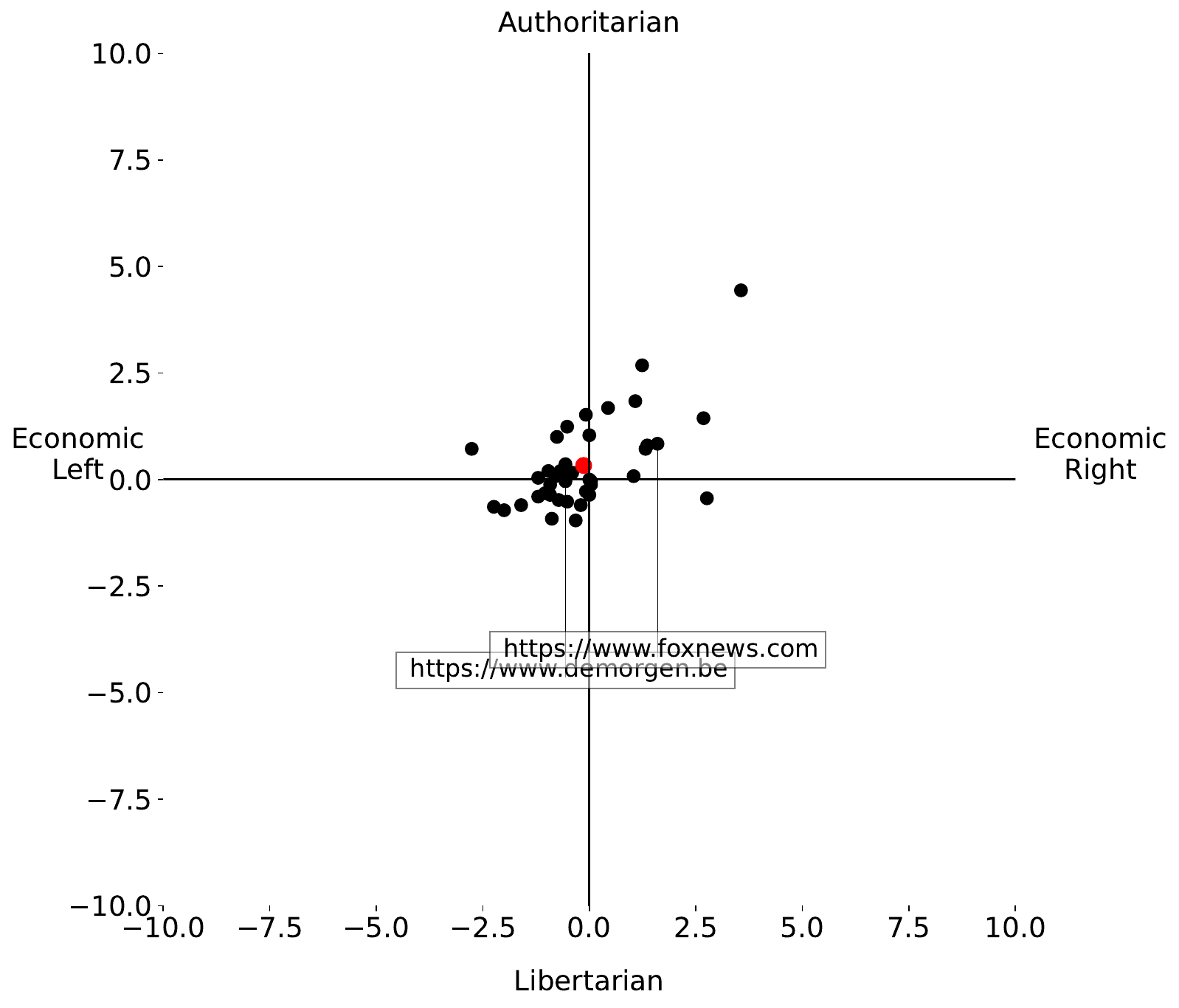}
        \caption{ChatGPT-4}
        \label{fig:sub1_2}
    \end{subfigure}
    
    \vspace{0.4cm} % add some vertical space between the rows

    \begin{subfigure}[b]{0.42\linewidth}
        \includegraphics[width=\linewidth]{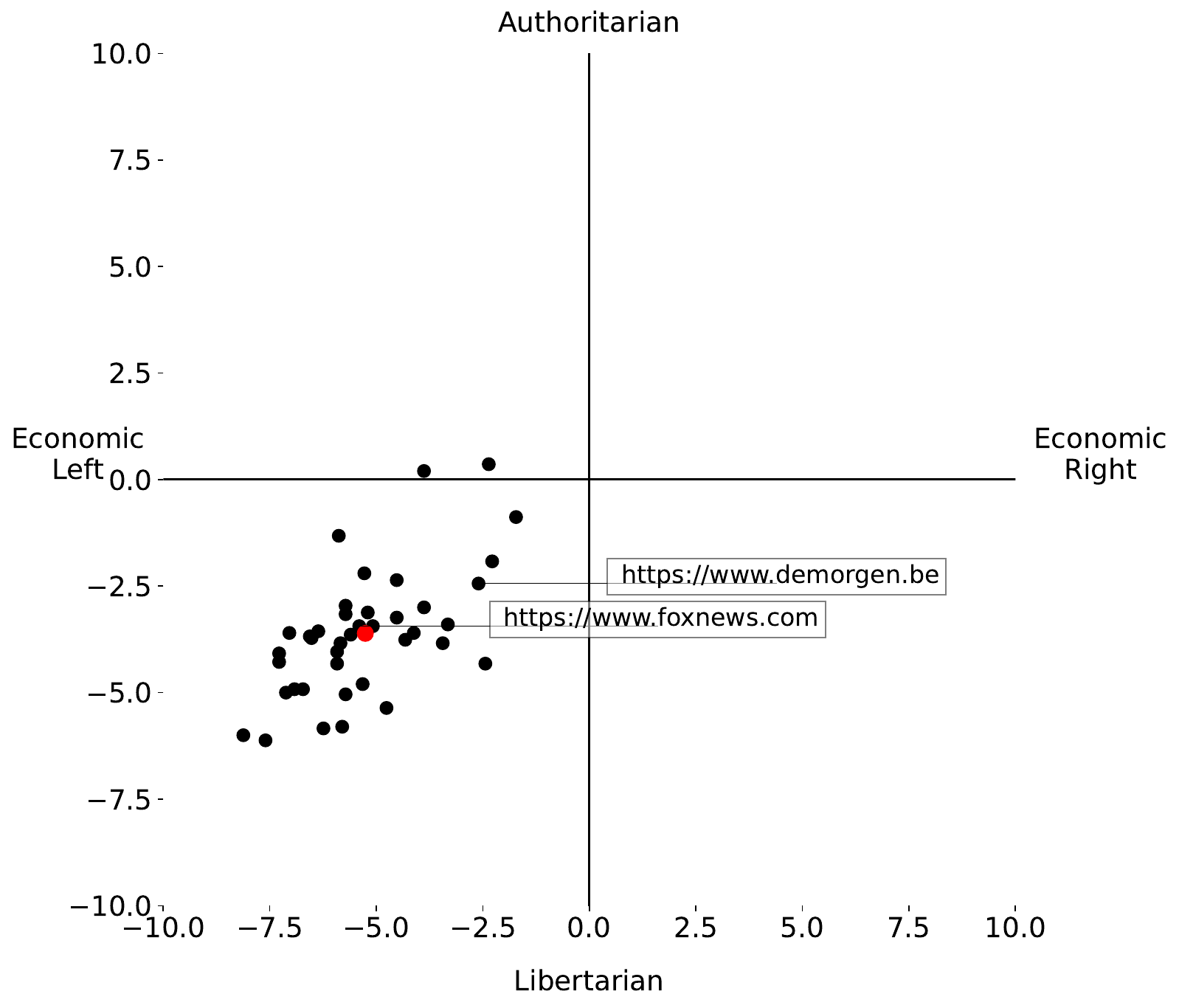}
        \caption{Gemini Pro}
        \label{fig:sub1_3}
    \end{subfigure}
    %\hfill
    \begin{subfigure}[b]{0.42\linewidth}
        \includegraphics[width=\linewidth]{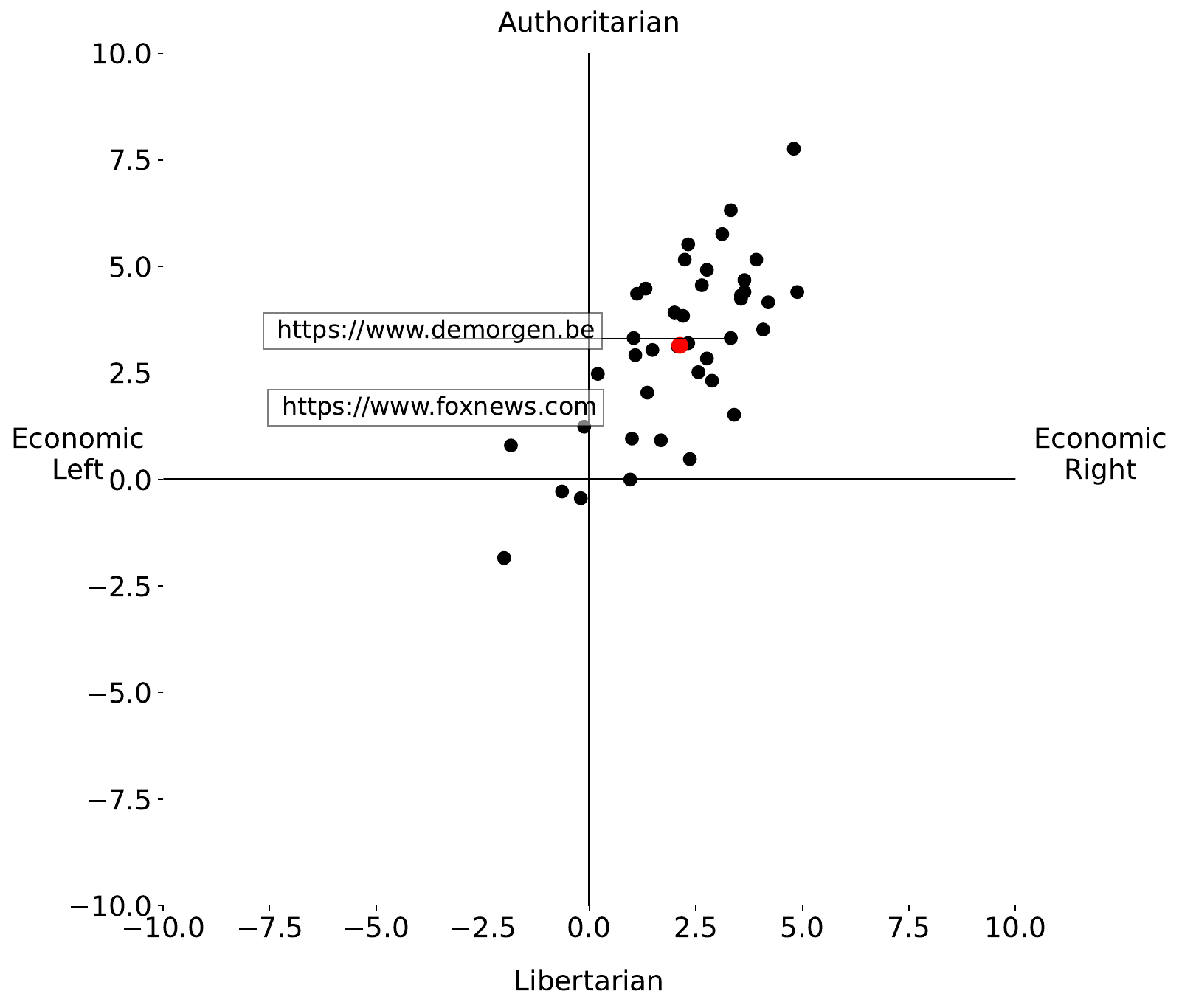}
        \caption{Gemini Pro 1.5}
        \label{fig:sub1_4}
    \end{subfigure}

    \caption{Distribution of newspapers positioning. Each black dot represents a newspaper, and is obtained by averaging the scores obtained over all articles from that newspaper. The red dot is the average over all newspapers. Two example newspapers (De Morgen and Fox News) are tracked throughout the four scatter plots, to exemplify how LLMs map them in the compass.}
    \label{fig:scatter}
\end{figure*}

To begin with, we have averaged the output of multiple articles from the same newspaper, to identify the mean positioning of each newspaper according to each LLM. \cref{fig:scatter} shows, for each LLM model, a scatter plot summarizing the mean evaluation for each of the 40 newspapers tested in our work. The red dots in the plots represent the mean value across all newspapers.

As evidenced in the plots, the four models seem to have largely different opinions on the political and economical viewpoints of the articles.
For ChatGPT-3.5 and Gemini Pro, the majority of the newspapers lay on the libertarian-left quadrant, with the latter model evaluating only two newspapers outside of this quadrant while the others are leaning towards the far-left. This does not match the presence of centre-right and right-winged newspapers in our dataset (\cref{tab:newspapers}).
By contrast, for Gemini Pro 1.5, the vast majority of newspapers tend to have an authoritarian and economically right stance, which is in contrast with the numerous group of centre-left and left-winged newspapers in the dataset. ChatGPT-4, on the contrary, groups almost all newspapers at the centre, but within a dense distribution that does not represent the broad political alignment contained in our dataset.

Overall, Gemini versions tend to be more extreme and have a large dispersion of points -- although they cannot agree on which quadrants of the compass --, whereas the ChatGPT versions cluster the newspapers on narrower distributions, closer to the centre (version 3.5) or centre-left (version 4). Impressively, there are no two LLMs giving the same distribution of news media positioning, but they all return their own judgement, without cross-consistency. In addition, no LLM seems to reproduce -- at least qualitatively -- the distribution in political alignment emerging from our dataset (\cref{tab:newspapers}). \\

\begin{figure*}[ht]
    \centering
    \begin{subfigure}[b]{0.42\linewidth}
        \includegraphics[width=\linewidth]{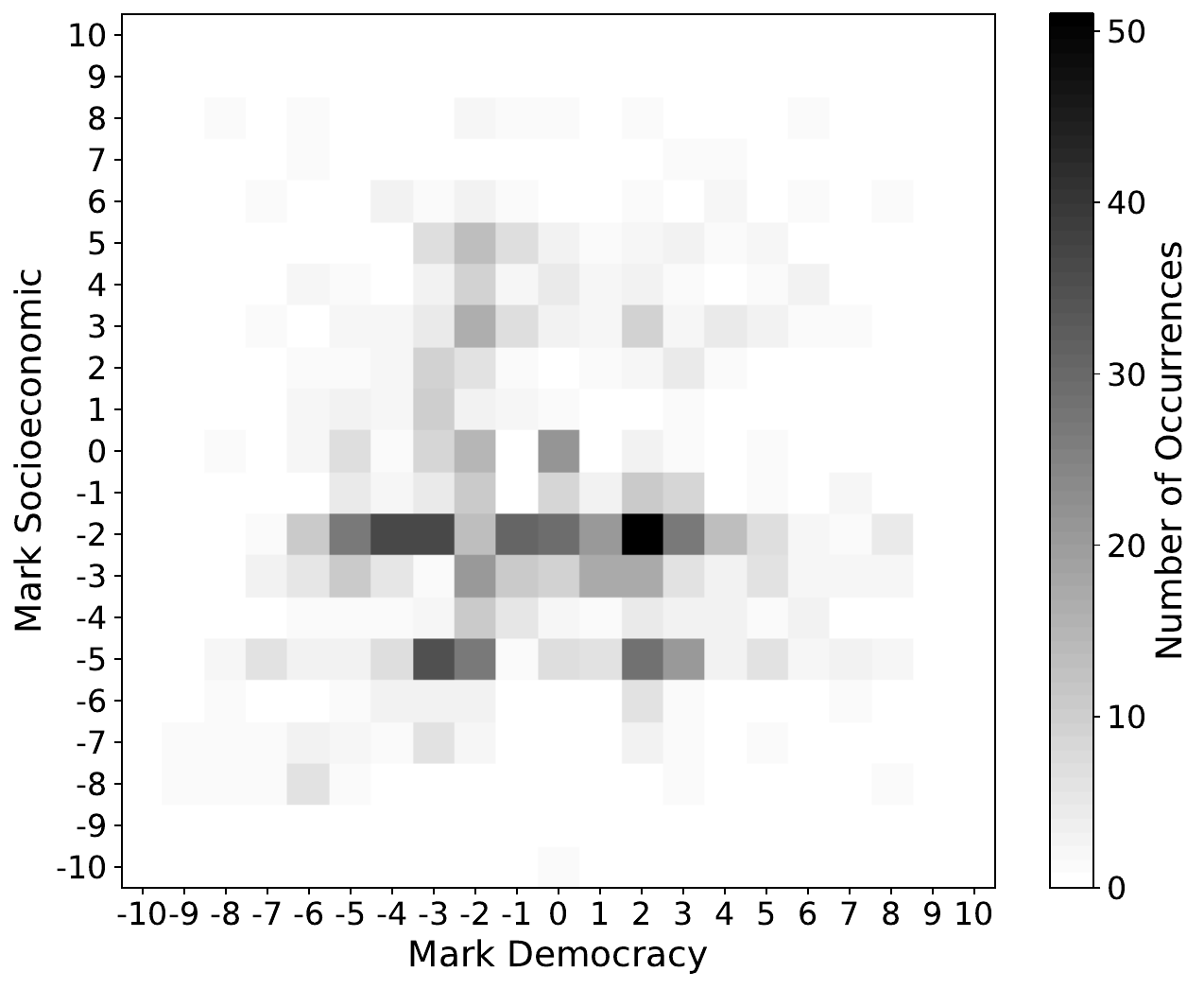}
        \caption{ChatGPT-3.5}
        \label{fig:sub2_1}
    \end{subfigure}
    %\hfill % this will add a little horizontal space between the subfigures
    \begin{subfigure}[b]{0.42\linewidth}
        \includegraphics[width=\linewidth]{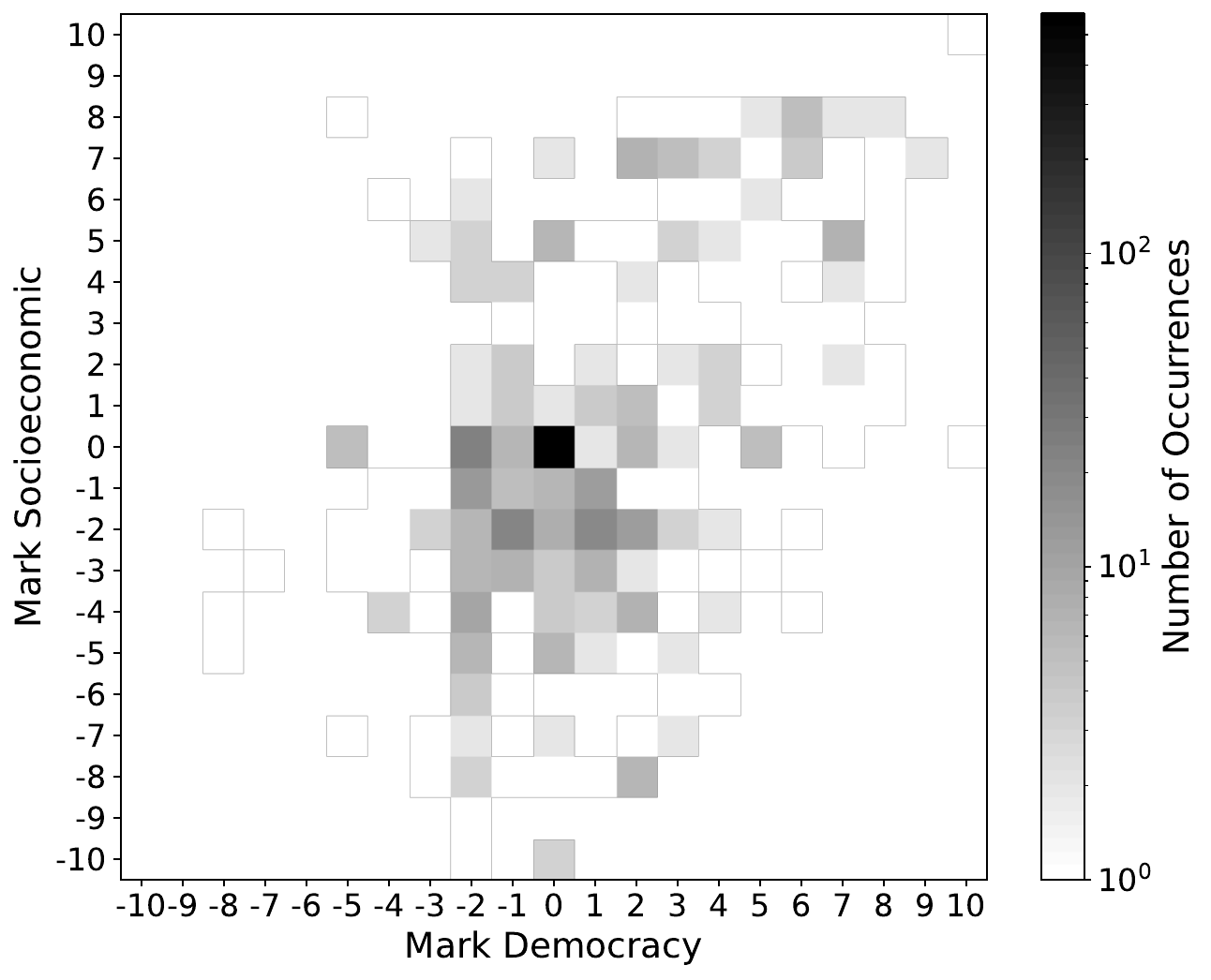}
        \caption{ChatGPT-4}
        \label{fig:sub2_2}
    \end{subfigure}
    
    \vspace{0.4cm} % add some vertical space between the rows

    \begin{subfigure}[b]{0.42\linewidth}
        \includegraphics[width=\linewidth]{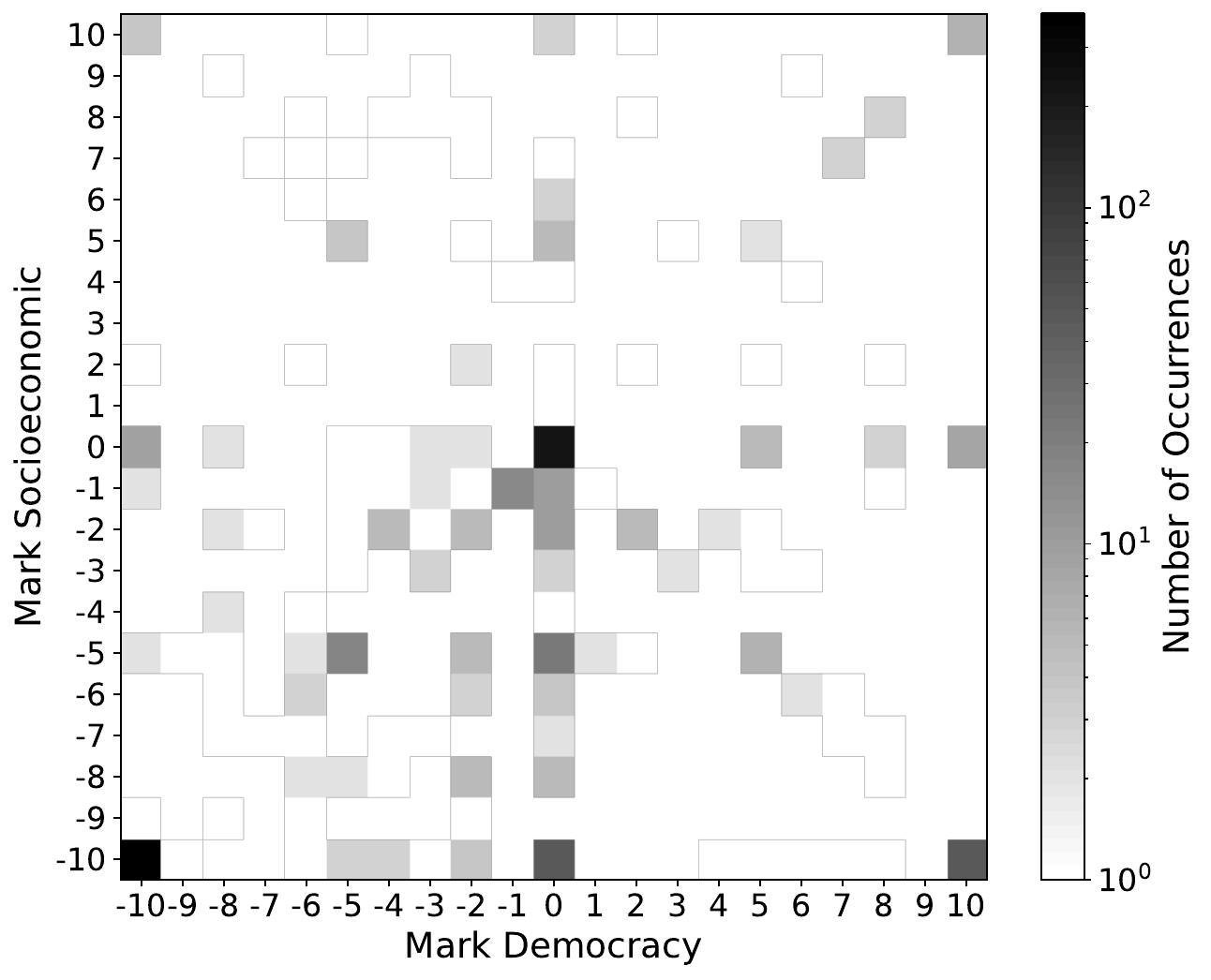}
        \caption{Gemini Pro}
        \label{fig:sub2_3}
    \end{subfigure}
    %\hfill
    \begin{subfigure}[b]{0.42\linewidth}
        \includegraphics[width=\linewidth]{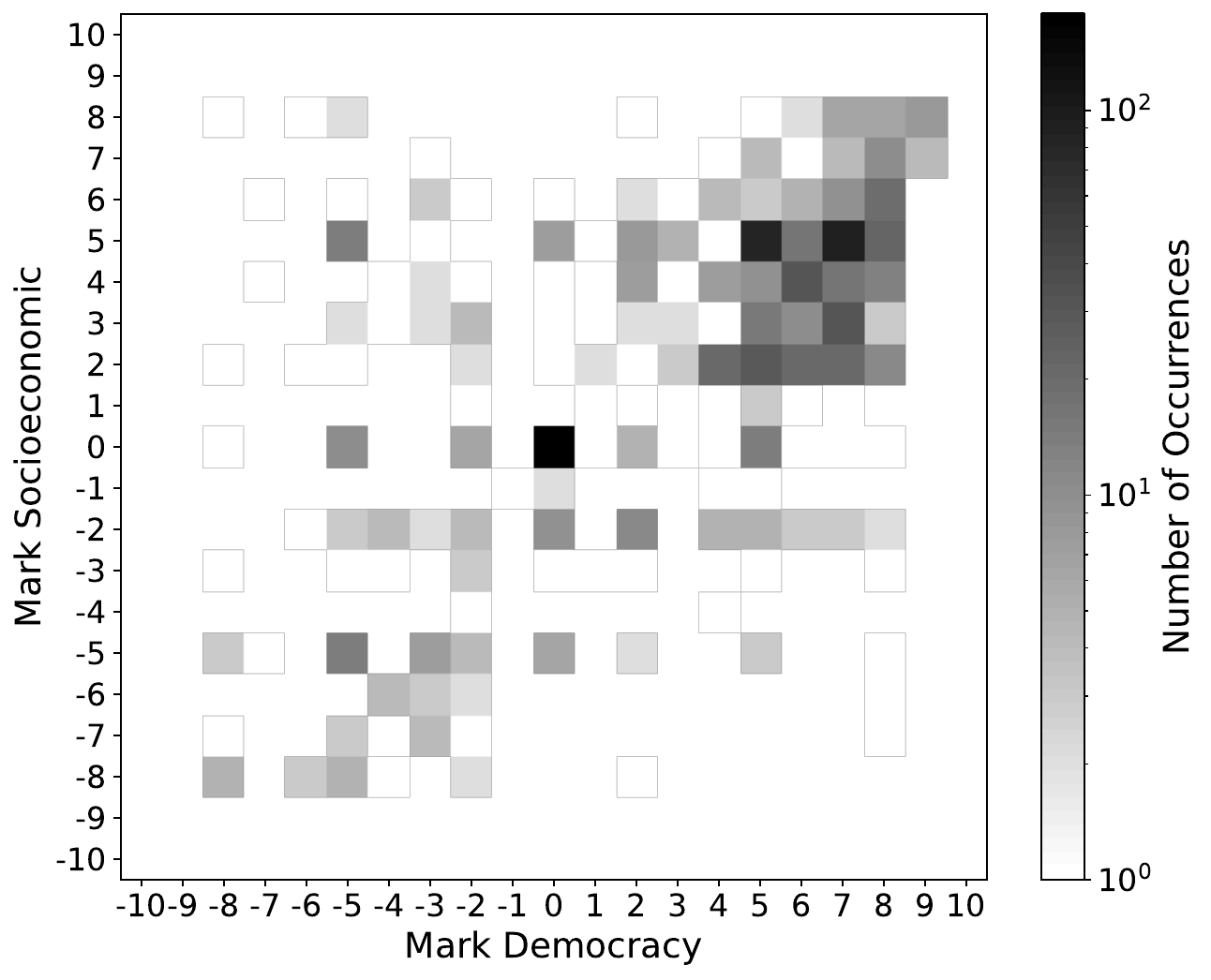}
        \caption{Gemini Pro 1.5}
        \label{fig:sub2_4}
    \end{subfigure}

    \caption{2D histograms (heatmap) of the positioning scores, over all articles. Each square represents a set of coordinates in the compass, and the color represents how many times such coordinate has been used by each LLM to map articles. Note that the ChatGPT-4 scale is logaritmic (\textit{cf.} Main Text).}
    \label{fig:heatmap}
\end{figure*}

To gain a clearer understanding of the results shown in \cref{fig:scatter}, and to make more sense of the seemingly randomicity of the responses, we present \cref{fig:heatmap}. It shows a heatmap for each model, which displays the frequency of each evaluating scores, over all articles (that is, how many times a certain score was used to locate articles). While \cref{fig:scatter} shows newspapers positioning, averaged over articles, \cref{fig:heatmap} investigates the rating of single articles. Note that, for ChatGPT-4, Gemini Pro 1.0 and Gemini Pro 1.5, we use a logarithmic scale on the number of occurrences, because a single evaluation accounts for over 100 occurrences -- which would render other evaluations much less visible on a linear scale.

As expected, the general leaning tendency observed in \cref{fig:scatter} is preserved in \cref{fig:heatmap}, confirming that LLMs are definitely not consistent with one another. Furthermore, the plot indicates that ChatGPT-3.5 exhibits the highest variance in its evaluations, covering much of the compass spectrum with a centre-left tendency. In contrast, ChatGPT-4 shows an overwhelmingly neutral judgment for these articles, with the vast majority of evaluations being [0,0] (recall the logaritmic scale). For Gemini Pro, most evaluations are concentrated at two points: [0,0] (perfect centrism) and [-10,-10] (far-left libertarian), with the latter accounting for 35\% of all cases. What seemed like a spread-out distribution towards centre-left in \cref{fig:scatter}, is instead a largely dichotomic judgement between centrism and far-left positioning. Gemini Pro 1.5 also issued many neutral judgments, albeit to a lesser extent than ChatGPT-4, plus a cluster positioned on the right-authoritarian quadrant. \\

\begin{figure*}[ht]
    \centering
    \begin{subfigure}[b]{0.48\linewidth}
        \includegraphics[width=\linewidth]{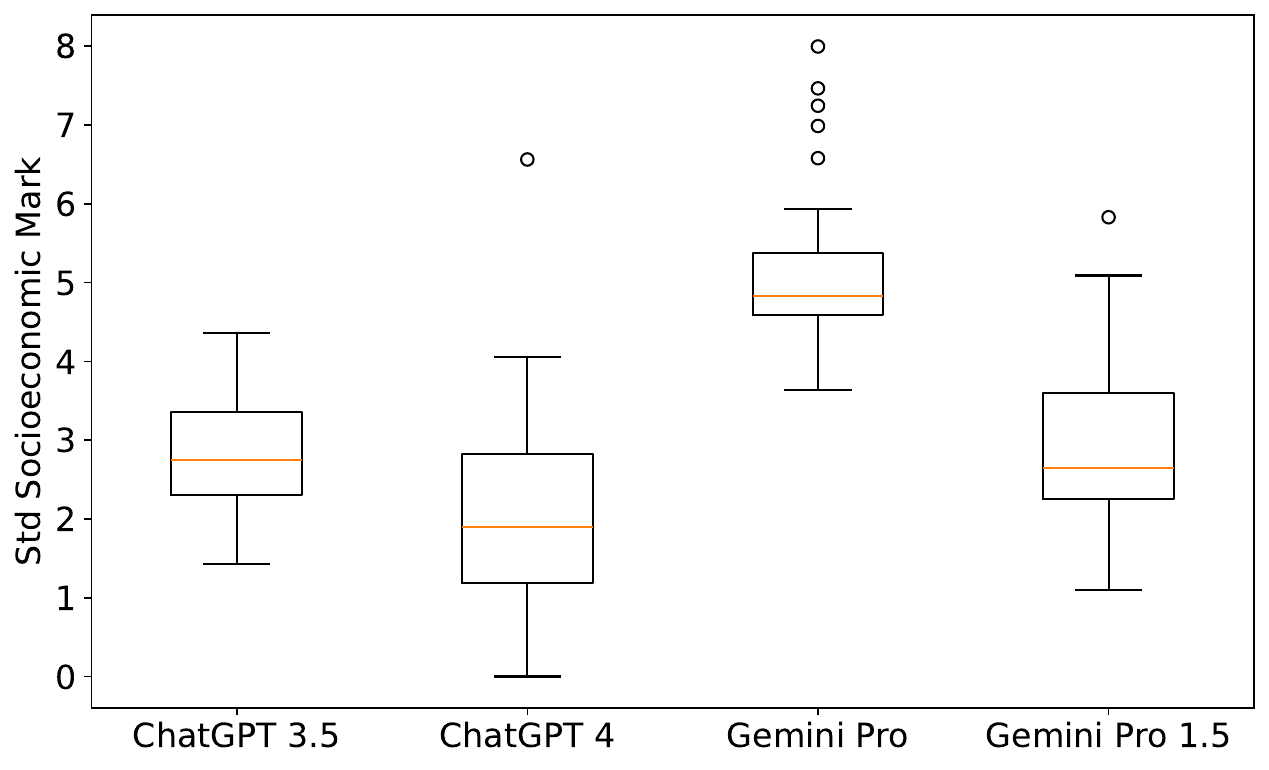}
        \label{fig:sub3_1}
    \end{subfigure}
    \hfill % this will add a little horizontal space between the subfigures
    \begin{subfigure}[b]{0.48\linewidth}
        \includegraphics[width=\linewidth]{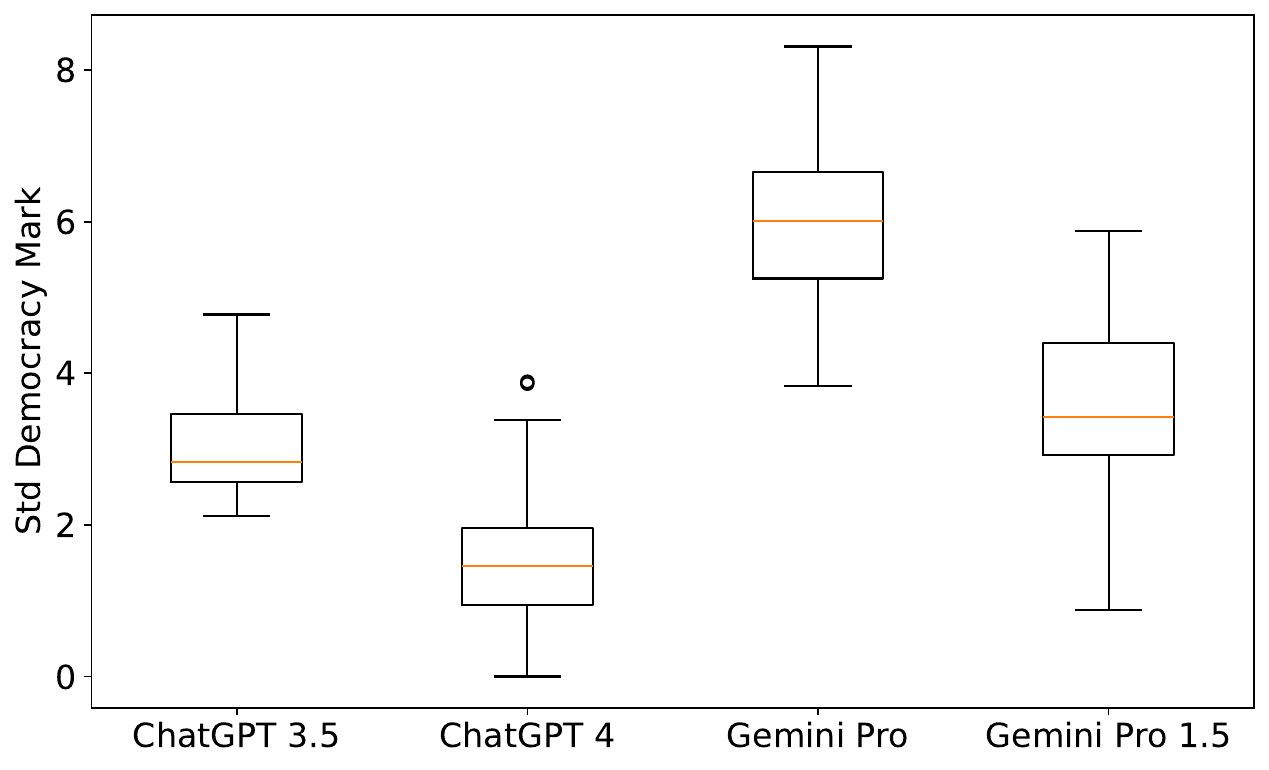}
        \label{fig:sub3_2}
    \end{subfigure}
    \caption{Boxplot distribution of standard deviations within each newspaper, for each considered LLM. The two panels refer to one dimension of evaluation -- socioeconomic or democracy marks.}
    \label{fig:boxplot}
\end{figure*}

Finally, we aimed at understanding not only how the evaluation of the selected LLMs varies across different newspapers, but also across articles within the same newspaper. Therefore, we calculated the standard deviation for each of the two dimensions —- socioeconomic and democracy —- across the articles of the same newspaper. \cref{fig:boxplot} presents the box plots (distributions) of the standard deviations across all newspapers.

For both dimensions, ChatGPT-4 has the lowest average standard deviation, confirming the insight about the dense clustering towards the centre. This can be likely attributed to the high percentage of [0,0] votes produced by the model, as discussed earlier. Then, the second most dense distribution comes from ChatGPT-3.5, followed by Gemini Pro 1.5 and Gemini Pro. Overall, the ChatGPT family seems to have less variability (even though, while evaluating the database from \cref{tab:newspapers}, that is rather spread out in the spectrum), while the Gemini family has wider distributions, even though one is leaning on opposing positions that the other, with left and right-winged biases, respectively. Given the high correlation between horizontal and vertical axis, the observations hold for both dimensions (\textit{cf.} the panels in \cref{fig:boxplot}).

Other notable observations include: 1) ChatGPT-3.5 has relatively narrow interquartile ranges for both dimensions, and 2) an absence of outliers for both dimensions; 3) Gemini Pro displays a relatively high number of outliers on the socioeconomic dimension. All in all, these observations confirm that LLM do not show any type of cross-consistency in their responses, not in the positioning scores nor in their statistics. 

\section{Discussion}
\label{sec:discussion}

Our results show a striking inconsistency across LLM responses when detecting the positioning of news media on an economical-political compass. They do not classify articles and newspapers similarly, nor they reproduce similar distributions of alignment. Overall, our study reveals that the models have significantly different opinions on the tested articles, suggesting the presence of errors and biases that, due to the inherent opacity of complex deep learning models, are extremely challenging to identify and correct.

On top of this quantitative investigation, we also recognise that, on a preliminary qualitative level, the responses are not even correctly aligned with expectations. In fact, our dataset includes newspapers in principle covering the full spectrum of orientations -- that is, according to the average tendency of each newspaper, as reported by various sources, see \cref{tab:newspapers}. Unfortunately, as noted in \cref{sub:methodology}, the 1,000 articles were not singularly reviewed and assessed by human experts. This makes it impossible, at this stage, to quantitatively evaluate each model's performance against a human ground truth.

% In this section, we discuss the necessity of human validation to advance research on this topic and examine the potential impact of LLMs in promoting fair journalism on a broader scale.

% \subsection{Human validation}
% \label{sub:benchmarking}

In this context, we thus make a call for actions to journalists and political scientists to engage in a comprehensive review of the selected articles. Human validation is vital for assessing the accuracy of judgments made by large language models in this field, and collective involvement is essential. Community efforts would thus significantly help not only to spot inconsistencies across LLMs, but also to verify the adequacy of their responses and to identify best practices or best-performing algorithms to set standards in this extremely delicate field. These studies will finally address both facets of whether LLMs work "properly".

The full list of tested articles is available in our GitHub repository \cite{githubRep}.
Additional articles can be evaluated using the same code used in this study, also found in the GitHub repository, or through our public non-profit platform, NAV AI \cite{navai}. On NAV AI, users can easily input the URL of an article and select a model for evaluation. Additionally, users can provide their own assessments, which will be anonymously collected, analysed and used as benchmarks.

\subsection{AI to reduce polarization in journalism}
\label{sub:reduce_polarization}

This study identified worrisome and sensitive shortcomings, to be promptly addressed and corrected by researchers and regulators. However, we remind that the field of AI, and more specifically LLMs, can help addressing the risks of polarization in journalism, and thus provide invaluable help for democracies. This can be achieved through multiple approaches:

\begin{itemize}
    \item \textbf{Media literacy}: Enhancing media literacy among the public can help individuals critically evaluate the information they encounter online. Education systems should incorporate curricula that teach critical thinking and fact-checking skills \cite{mihailidis2017spreadable, guess2020digital}. LLMs can be used in this regard to offer interactive learning tools, assist with fact checking, recommend resources and support educators in making their teachings more effective. 
    \item \textbf{Support for quality journalism}: Investing in quality journalism, particularly at the local level, is crucial. Public funding, grants, and subscriptions can help sustain independent news organizations \cite{pickard2020democracy}.
    LLMs can be used in this regard with fact-checking, the automation of routine tasks like transcriptions, translation, formatting etc., by providing advanced data analytics, and by helping to write grant proposals.
    \item \textbf{Regulation and oversight}: Governments and international bodies should develop regulations to combat misinformation and hold social media platforms accountable for the content they disseminate. This regulation must balance the need to curb harmful content with the protection of free speech \cite{gorwa2019platform}.
    LLMs can be employed to continuously monitor social media platforms for misinformation. They can analyze vast amounts of content in real-time, flagging potentially harmful or misleading information for further review by human moderators. LLMs can be used to analyze trends in misinformation, identifying common themes, sources, and methods of dissemination. The results may inform regulatory bodies and social media platforms about emerging threats and help in developing targeted countermeasures. 
    LLMs can also assist researchers and policy-makers in compiling reports and studies on the effectiveness of various regulatory measures, and in developing balanced guidelines. By analyzing legal precedents, human rights frameworks, and public sentiment, LLMs can help craft nuanced policies that respect free expression while addressing misinformation.
\end{itemize}

We hope that, by following our study, instruments of this kind can be developed, validated and shared.

\subsection{Future work}
\label{sec:future_work}

In this study, we have conducted an extensive comparison between OpenAI's ChatGPT series and Google's Gemini series, aiming to elucidate the similarities and differences in how these models assess the political spectrum of newspapers worldwide.
Future works may include:

\begin{enumerate}
    \item Incorporating and evaluating more models, particularly local ones such as Llama3, to broaden the scope and depth of our analyses.
    \item Benchmark the evaluation of the models against human ground truth evaluations (see \cref{sec:discussion}), in a multidisciplinary endeavour within the digital humanities field.
    \item If these models are deemed reliable through human benchmarking, future research should focus on evaluating a broader dataset of newspaper articles. Of particular benefit would be evaluating articles from the same newspapers but spanning multiple years. This would enable the exploration of potential shifts in the average political opinions expressed in these newspapers. 
    \item Such systematic dynamical analysis would unravel the trajectories of news media and the degree of polarization towards extreme ideologies, as well as any trends towards more authoritarian stances. 
    This expanded analysis will provide deeper insights into the dynamics of media bias and the evolution of journalistic tones over time, contributing significantly to the understanding of media influence on public perception and discourse.
\end{enumerate}

\section{Conclusion}
\label{sec:conclusion}

The digital age has revolutionized journalism, providing unparalleled opportunities for information dissemination and public engagement. However, it has also brought significant risks, including the polarization of ideas and the spread of misinformation.

In this study, we explored whether state-of-the-art LLMs can assist in identifying the political and economical orientation of newspapers. We tested ChatGPT-3.5, ChatGPT-4, Gemini Pro and Gemini Pro 1.5 by evaluating a set of 1,000 articles from 40 newspapers across 27 countries using a two-dimensional scale: socioeconomic (left to right) and democratic (libertarian to authoritarian) \cite{heywood2021political}. The results revealed a high level of inconsistencies among the models, with estimated opinions scattered all over the compass, differing across LLMs and hardly reproducing what was expected from the database.
Additionally, the models exhibited varying levels of randomness in their evaluations of individual papers. ChatGPT-4 was the less volatile LLM, while Gemini Pro was the most variable one. All these findings advocate for deeper investigations of LLMs -- including data collection, development and training -- in order to reduce biases and poor results, which may greatly impact sensitive endeavours such as fact-checking and news media monitoring.

We recall that, by this study, we have addressed the first question in LLM evaluation in the digital humanities, that is, whether they provide consistent responses with one another. A necessary next step would be to investigate the second point, i.e., quantitatively evaluating each model's performance against a human benchmark. On top of our preliminary analysis in this direction, we have issued a call to action for experts to contribute to our future research by evaluating the political stances of the articles in this study (e.g. using the code available at \cite{githubRep}) and listing and evaluating new articles on our public platform NAV AI (\url{https://navai.pro/}) \cite{navai}.

Lastly, we advocate for the use of LLMs to mitigate the risks of polarization by promoting media literacy, supporting quality journalism, and implementing effective regulations. We believe that AI can help ensure that journalism continues to fulfill its vital role in democratic societies -- but, for that, we need better AI than today.

\printbibliography

\newpage

\onecolumn
\begin{appendices}

\section{Eligible newspapers for test set}
\label{app:newspapers}

\cref{tab:all_newspapers} reports the complete list of newspapers that were initially considered for our investigation.

\begin{table*}[htbp]
\centering
\begin{tabular}{ll|ll|ll}
Newspaper & Country & Newspaper & Country & Newspaper & Country \\
\hline
Clarín & ARG & Clarín (Argentina) & ARG & La Nación & ARG \\
Kronen Zeitung & AUT & Kurier & AUT & Wiener Zeitung & AUT \\
The Australian & AUS & Sydney Morning Herald & AUS & De Morgen & BEL \\
Le Soir & BEL & De Standaard & BEL & The Rio Times & BRA \\
O Globo & BRA & Folha de S.Paulo & BRA & O Globo & BRA \\
O Estado de S. Paulo & BRA & Toronto Star & CAN & The Globe and Mail & CAN \\
The Toronto Star & CAN & Le Devoir & CAN & The Globe and Mail & CAN \\
Toronto Sun & CAN & La Presse & CAN & Le Nouvelliste & CHE \\
El Mercurio & CHL & Guangzhou Daily & CHN & People's Daily & CHN \\
China Daily & CHN & El Tiempo & COL & Lidové noviny & CZE \\
Frankfurter Allgemeine Zeitung & DEU & Die Welt & DEU & Die Zeit & DEU \\
Bild & DEU & Der Tagesspiegel & DEU & Ekstra Bladet & DNK \\
Berlingske & DNK & Politiken & DNK & Postimees & EST \\
El Mundo & ESP & La Vanguardia & ESP & El País & ESP \\
La Razón & ESP & Helsingin Sanomat & FIN & Libération & FRA \\
Le Monde & FRA & La Tribune & FRA & Le Figaro & FRA \\
Kathimerini & GRC & The Standard & HKG & 168 Óra & HUN \\
Alfahír & HUN & About Hungary & HUN & 24.hu & HUN \\
Kompas & IDN & The Jakarta Post & IDN & Koran Tempo & IDN \\
Jakarta Post & IDN & The Irish Independent & IRL & The Irish Times & IRL \\
The Jerusalem Post & ISR & The Times of Israel & ISR & The Hindu & IND \\
Rajasthan Patrika & IND & Malayala Manorama & IND & Amar Ujala & IND \\
Business Standard & IND & Dainik Jagran & IND & Bhaskar & IND \\
The Hindu & IND & The Times of India & IND & Mint & IND \\
Il Sole 24 Ore & ITA & La Repubblica & ITA & Il Corriere della Sera & ITA \\
Libero & ITA & Ad-Dustour & JOR & The Mainichi Newspapers & JPN \\
The Nikkei & JPN & The Chunichi Shimbun & JPN & The Yomiuri Shimbun & JPN \\
The Asahi Shimbun & JPN & Daily Nation & KEN & The Korea Herald & KOR \\
The Korea Times & KOR & The Chosun Ilbo & KOR & The Daily Star & LBN \\
Luxemburger Wort & LUX & Le Quotidien & LUX & Tageblatt & LUX \\
El Universal & MEX & La Jornada & MEX & Excelsior & MEX \\
The Star & MYS & Malaysia Kini & MYS & The Punch & NGA \\
Algemeen Dagblad & NLD & NRC Handelsblad & NLD & de Volkskrant & NLD \\
Aftenposten & NOR & The New Zealand Herald & NZL & New Zealand Herald & NZL \\
The New Zealand Herald & NZL & El Comercio (Peru) & PER & Dawn & PAK \\
Gazeta Wyborcza & POL & Correio da Manhã & PRT & Jornal de Notícias & PRT \\
Público & PRT & A Bola & PRT & Al-Watan & QAT \\
Al Sharq & QAT & Al Jazeera & QAT & Komsomolskaya Pravda & RUS \\
Kommersant & RUS & Rossiyskaya Gazeta & RUS & Aftonbladet & SWE \\
Dagens Nyheter & SWE & The Straits Times & SGP & Pravda & SVK \\
Bangkok Post & THA & The Nation (Thailand) & THA & Bangkok Post & THA \\
Hürriyet & TUR & Cumhuriyet & TUR & China Post & TWN \\
The Guardian & GBR & The Daily Mail & GBR & The Telegraph & GBR \\
The Times & GBR & BBC & GBR & The Scotsman & GBR \\
The Independent & GBR & Daily Mirror & GBR & The Atlanta Journal-Constitution & USA \\
The Washington Post & USA & CNN & USA & Fox News & USA \\
Forbes & USA & The Philadelphia Inquirer & USA & Chicago Tribune & USA \\
Los Angeles Times & USA & The Boston Globe & USA & San Francisco Chronicle & USA \\
Star Tribune & USA & The Oregonian & USA & The New York Times & USA \\
El País & URY & The Cape Times & ZAF & Sunday Times & ZAF \\
The Sowetan & ZAF & Die Burger & ZAF &  &  \\
\end{tabular}
\caption{Initial list of newspapers}
\label{tab:all_newspapers}
\end{table*}

\end{appendices}

\end{document}